\documentclass[conference]{IEEEtran}
\IEEEoverridecommandlockouts
\usepackage{cite}
\usepackage{amsmath,amssymb,amsfonts}
\usepackage{graphicx}
\usepackage{textcomp}
\usepackage{xcolor}
\usepackage{subcaption}
\usepackage{algorithm}
\usepackage{algpseudocode}
\usepackage{longtable}
\usepackage{soul}
\usepackage{pifont}
\usepackage{bm}
\usepackage{booktabs}
\usepackage{multicol}
\usepackage{multirow}
\usepackage{array}
\usepackage{url}
\usepackage{stfloats}
\usepackage{soul}
\usepackage{arydshln}

\usepackage{xspace}
\usepackage{verbatim}
\usepackage{microtype}
\usepackage{enumitem}
\usepackage{bbold}
\usepackage{newfloat}

\newcommand{\cmark}{$\surd$}%
\newcommand{\xmark}{$\times$}%

\def\BibTeX{{\rm B\kern-.05em{\sc i\kern-.025em b}\kern-.08em
    T\kern-.1667em\lower.7ex\hbox{E}\kern-.125emX}}
\begin{document}

\title{Contextual Explainable Video Representation:\\Human Perception-based Understanding
}

\author{\IEEEauthorblockN{Khoa Vo}
\IEEEauthorblockA{\textit{Dept. of CSCE} \\
\textit{University of Arkansas}\\
Fayetteville, AR, USA\\
khoavoho@uark.edu}
\\
\IEEEauthorblockN{Phat Nguyen }
\IEEEauthorblockA{\textit{AI Lab} \\
\textit{FPT Software}\\
Ho Chi Minh City, Vietnam\\
phongnx1@fsoft.com.vn}
\and
\IEEEauthorblockN{Kashu Yamazaki}
\IEEEauthorblockA{\textit{Depth. of CSCE} \\
\textit{University of Arkansas}\\
Fayetteville, AR, USA\\
kyamazak@uark.edu}
\\
\IEEEauthorblockN{Khoa Luu}
\IEEEauthorblockA{\textit{Dept. of CSCE} \\
\textit{University of Arkansas}\\
Fayetteville, AR, USA\\
khoaluu@uark.edu}
\and
\IEEEauthorblockN{Phong X. Nguyen}
\IEEEauthorblockA{\textit{AI Lab} \\
\textit{FPT Software}\\
Ho Chi Minh City, Vietnam\\
phatnt21@fsoft.com.vn}
\\
\IEEEauthorblockN{Ngan Le}
\IEEEauthorblockA{\textit{Dept. of CSCE} \\
\textit{University of Arkansas}\\
Fayetteville, AR, USA\\
thile@uark.edu}
}

\maketitle

\begin{abstract}
Video understanding is a growing field and a subject of intense research, which includes many interesting tasks to understanding both spatial and temporal information, e.g., action detection, action recognition, video captioning, video retrieval. One of the most challenging problems in video understanding is dealing with feature extraction, i.e. extract contextual visual representation from given untrimmed video due to the long and complicated temporal structure of unconstrained videos. Different from existing approaches, which apply a pre-trained backbone network as a black-box to extract visual representation, our approach aims to extract the most contextual information with an explainable mechanism. As we observed, humans typically perceive a video through the interactions between three main factors, i.e., the actors, the relevant objects, and the surrounding environment. Therefore, it is very crucial to design a contextual explainable video representation extraction that can capture each of such factors and model the relationships between them. In this paper, we discuss approaches, that incorporate the human perception process into modeling actors, objects, and the environment. We choose video paragraph captioning and temporal action detection to illustrate the effectiveness of human perception based-contextual representation in video understanding. Source code is publicly available at \url{https://github.com/UARK-AICV/Video_Representation}.
\end{abstract}


\section{Introduction}
\label{sec:intro}

Video understanding is one of the fundamental field in computer vision that comprises of a wide range of tasks that deal with datasets of videos. These tasks commonly require to extract essential information from the input videos in order to serve different goals.

Based on the present of video pre-processing, we can divide video understanding tasks into two categories of trimmed videos tasks and untrimmed videos tasks. On the one hand, tasks on trimmed videos such as action recognition \cite{pareek2021survey, vu2021teaching, wang2021tdn, vu2021self, sun2022human, vu20222+} or video captioning require input videos to be perfectly trimmed to contain no irrelevant frames (e.g., background frames). On the other hand, tasks on untrimmed videos such as temporal action proposals generation (TAPG)~\cite{lin2018bsn, bmn, dbg, KhoaVo_ICASSP, KhoaVo_Access, khoavo_aei_bmvc21, vo2022aoe}, temporal action detection (TAD)~\cite{xu2020gtad, pgcn_cvpr2020, zhang2022actionformer}, , video paragraph captioning (VPC)~\cite{lei2020mart, dai2019transformer, kashu_VLTinT}, video retrieval \cite{snoek2009concept, gabeur2020multi, wang2021t2vlad, wray2021semantic}, etc. can process on arbitrary untrimmed videos. In this paper, we focus on the tasks on untrimmed videos not only because they are more challenging in dealing with uncleaned videos but also because they are fundamental tasks to automatically trim the videos or extract crucial information and eliminate irrelevant segments. Particularly, we will provide details discussion of TAPG and VPC as specific tasks. 

Given an untrimmed video, TAPG requires to localize intervals for each presenting action or activity of interest. TAPG is a fundamental task for various downsteam applications, e.g., TAD and VPC. More specifically, TAD additionally requires an action label along with every proposed interval. On VPC, the intervals extracted by TAPG are jointly used to generate a coherent paragraph that describes important events of the input video.

\begin{figure}[t]
    \centering
    \includegraphics[width=0.45\textwidth]{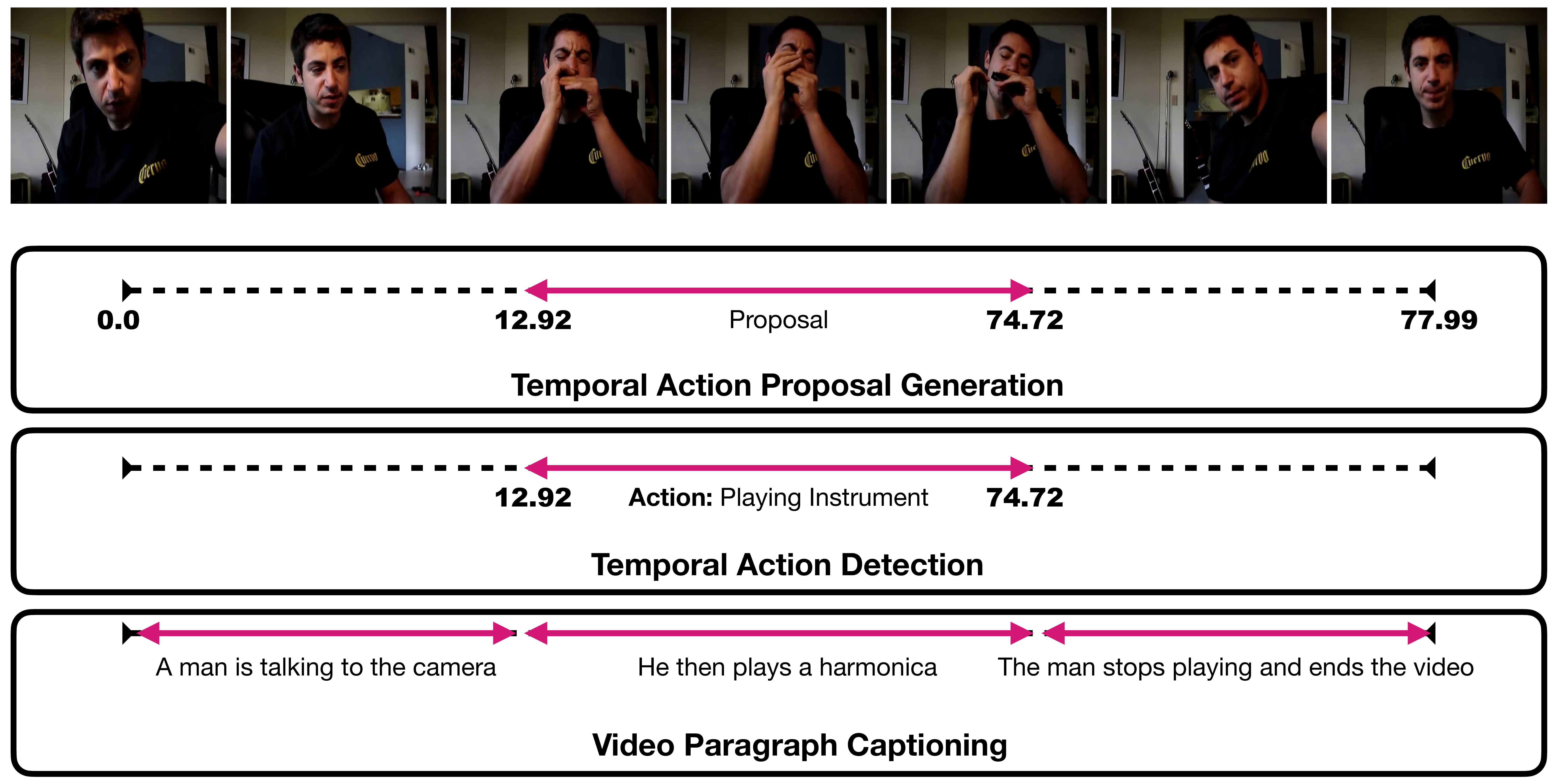}
    \caption{An illustration of tasks on an untrimmed video, including temporal action proposals generation (top box), temporal action detection (middle box), and video paragraph captioning (bottom box).}
    \label{fig:tasks}
    \vspace{-5mm}
\end{figure}

\begin{figure}
  \centering
  \begin{subfigure}[b]{0.45\textwidth}
    \centering
    \includegraphics[width=\textwidth]{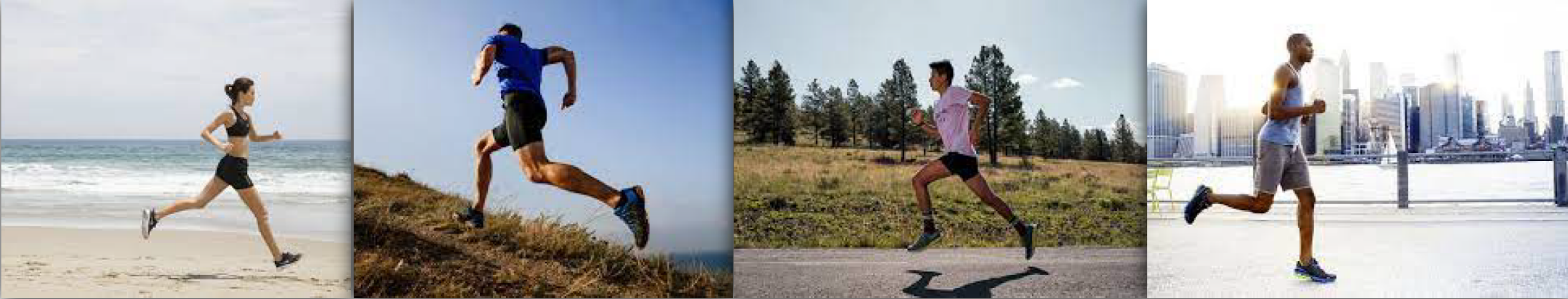}
    \caption{Examples of actions (e.g jogging) are independent to environments.}
    \label{fig:motivation1}
  \end{subfigure}
  \hfill
  \begin{subfigure}[b]{0.45\textwidth}
    \centering
    \includegraphics[width=\textwidth]{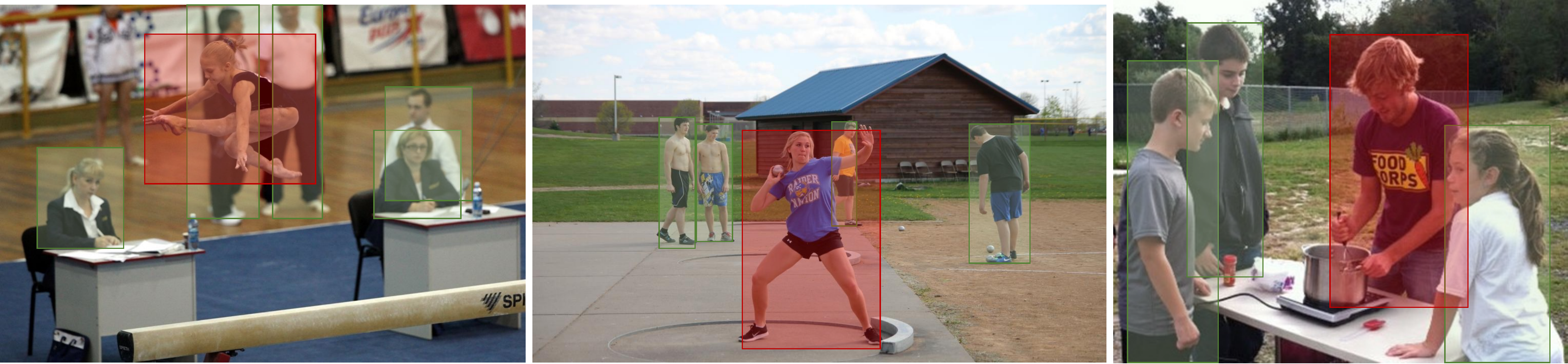}
    \caption{Examples of how actors contribute to form actions i.e. among all actors (green and red boxes) in the scenes, only main actors (red boxes) actually commit actions.}
    \label{fig:motivation2}
  \end{subfigure}
  \hfill
  \begin{subfigure}[b]{0.45\textwidth}
    \centering
    \includegraphics[width=\textwidth]{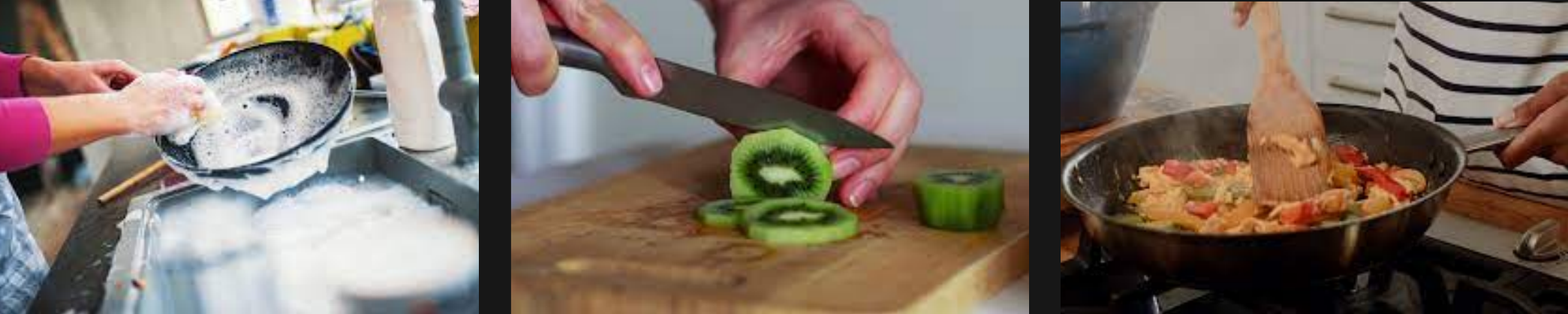}
    \caption{Examples of actions in egocentric videos where actors are not visible.}
    \label{fig:motivation3}
  \end{subfigure}
     \hfill
     \begin{subfigure}[b]{0.45\textwidth}
         \centering
         \includegraphics[width=\textwidth]{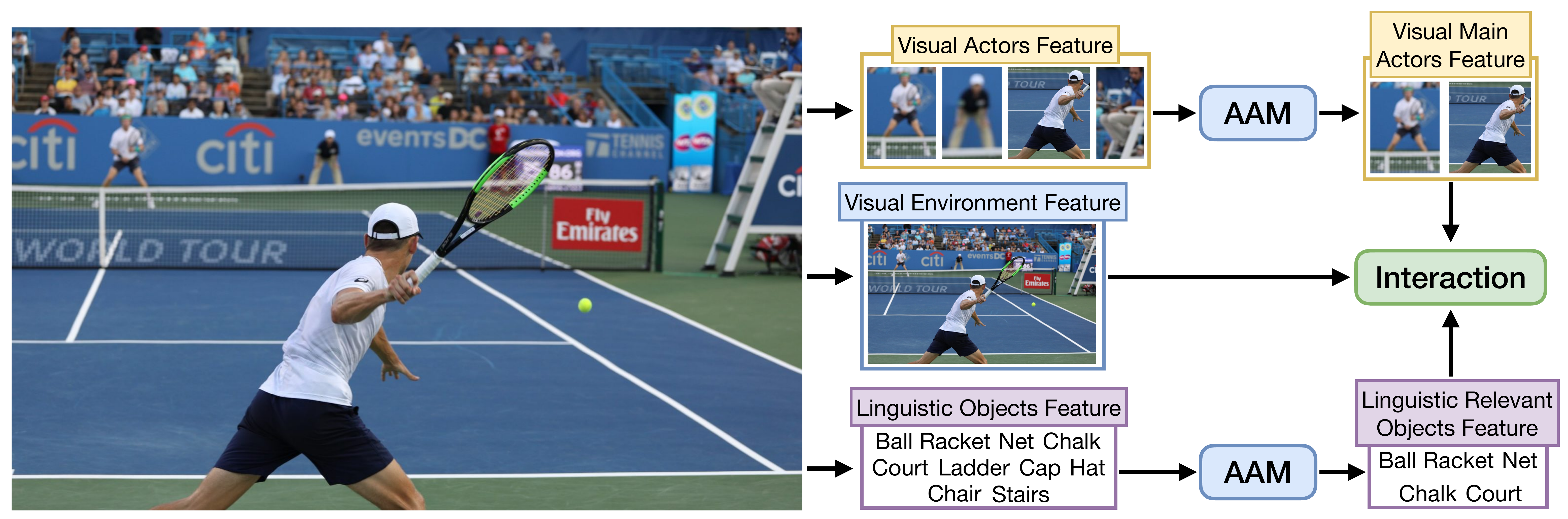}
         \caption{Our proposed Perception-based Representation (PMR) is modeled by both global visual environment, local visual main actors features, linguistic relevant objects features, and the interaction among them. In PMR, our proposed Adaptive Attention Mechanism (AAM) is to select main actors and relevant objects.}
         \label{fig:ourAOE}
     \end{subfigure}
  \caption{Most existing TAPG methods \cite{lin2018bsn, BSN++, bmn, dbg, xu2020gtad} apply a 3D backbone network to entire spatial domain. However, as shown in (a), actors contribute more importance to an action than environment itself. Moreover, (b) shows that main actors who actually commit actions may be among many inessential actors, or (c) actors are not visible in the scene of egocentric videos. \textit{This figure is cited from \cite{vo2022aoe}}.}
  \label{fig:motivation}
  \vspace{-5mm}
\end{figure}

Although TAPG and VPC methods have made great progresses in popular benchmarks of ActivityNet-1.3 \cite{caba2015activitynet}, THUMOS-14 \cite{THUMOS14}, or ActivityNet Captions \cite{krishna2017dense}, they still possess a common limitation, which is the overlooked video representation. In the respective module of TAPG and VPC methods, the input video frames are clustered into snippets of $\delta$ frames, then, a pre-trained 3D convolutional network \cite{c3d, i3d_2017} is used to encode each snippet to a feature vector. Despite being pre-trained on a large dataset (e.g., Kinetics \cite{Kinetics}) and able to compress semantic and movement information of the entire snippet in just a feature vector, such feature easily misses information from humans or objects appearing in smaller regions and tends to be biased to the overall spatial environment. Such neglected video representation leads to weak representations for hard scenarios illustrated in Fig.~\ref{fig:motivation}. Those scenarios can be briefly described as follows:
\begin{itemize}
    \item Scenario 1: Existing visual representation is easily biased by environment whereas the action may be independent to the environment as shown in Fig.~\ref{fig:motivation1}. This becomes more problematic when the actors occupy smaller regions compared to the overall environment.
    \item Scenario 2: An arbitrary number of actors can appear in the scene at the same time, but only a few of them are main actors that actually contribute to the formation of an action.
    \item Scenario 3: The main actor may not even appear inside video frames but only shows their hands interacting with objects to perform actions.
\end{itemize}

Furthermore, understanding a video involves multiple factors such as single human actor, group human actors, non-human actor, phenomenon \cite{KhoaVo_Access, KhoaVo_ICASSP, khoavo_aei_bmvc21, vo2022aoe}. Examples of non-human actors and phenomena performing actions include dog chasing, car running, and cloud floating.

Inspired by how humans perceives a video (i.e., at a specific timestamp, a human would look at overall scene, then localizing main actors, and perceiving objects that they interact with), our Perception-based Multi-modal Representation (PMR) is proposed in order to comprehensively capture crucial information from multiple entities in the spatial scene of each input snippet of the video. In order to do that, PMR consists of four modules: (i) Environment Beholder, which models the overall scene of input snippet, (ii) Actors Beholder, which models main actors appearing in the input snippet, (iii) Objects Beholder, which models relevant objects of the snippet, and (iv) Actors-Objects-Environment Beholder, which models the relationships between all types of entities. Furthermore, Actors Beholder and Objects Beholder are equipped with our newly proposed Adaptive Attention Mechanism (AAM) to eliminate inessential actors and irrelevant objects, respectively, appearing in the scene and only apply self-attention mechanism on main actors and most relevant objects, respectively.

Our contribution can be summarized as follows:
\begin{itemize}
    \item A discussion about our proposed Multi-modal Representation (PMR) that comprehensively represent video snippets.
    \item The integration of PMR with state-of-the-art (SOTA) methods in various tasks on untrimmed videos, including TAPG and VPC.
    \item Extensive experiments showing the effectiveness of PMR in the above tasks by creating a large performance margin over existing SOTAs.
\end{itemize}

\section{"Graybox" Contextual Explainable Representation: A journey}
\label{sec:graybox}
In this section, we address all aforementioned limitations by introducing a journey of developing a "graybox" contextual explainable representation. Our journey is step-by-step introduced as follows:

\subsection{Actors - Environment Interaction}

To alleviate Limitation 1 stated in Sec.~\ref{sec:intro},  we propose to model each snippet by two separate entities of local actors and global surrounding environment in \cite{KhoaVo_ICASSP, KhoaVo_Access}. For global environment, we extracted a feature map of the snippet by a pre-trained 3D convolutional network \cite{c3d, TSN2016ECCV} and apply average pooling on the feature map to obtain a single feature vector representing the environment. For local actors, we use an off-the-shelf human detector to localize them using the middle frame of the snippet, each detected bounding box is aligned onto the feature map extracted during the global environment processing to form a set of features for all actors, which in turns are fused together into a single actors feature using a self-attention module \cite{attention_is_all_you_need}. Both features of actors and environment are combined by another self-attention module to flexibly balance between local and global visual representation.

\subsection{Main Actors - Environment Interaction}
Limitation 2 poses a very common case where many actors appear in the scene but only several of them are main actors who actually contribute to the actions of interest. To resolve such case, we propose an adaptive attention mechanism (AAM) \cite{khoavo_aei_bmvc21}, which aims to (i) eliminate inessential actors who do not majorly affect the content of the scene and can be treated as background, and (ii) adaptively fuse information of selected main actors into a single feature vector. 

\begin{algorithm}[t]
\caption{AAM to extract the representation of main actors in a snippet.}
\label{algo:aam}
\hrule
\begin{algorithmic}[1]
\algrenewcommand\algorithmicrequire{\textbf{Data: }}
\algrenewcommand\algorithmicensure{\textbf{Result: }}
\Require Feature vector $f^e$ and features set $\mathcal{F}^a$ represent environment and all actors that appear in an input snippet, respectively.
\Ensure Feature vector $f^a$ represents main actors.

\State $\hat{f}^e \gets MLP_{\theta_e}(f^e)$ 
\State set $\tilde{\mathcal{F}}^a$, $H^a$ to empty list \Comment{\scriptsize{$\mathcal{F}^a$ stores selected main actors, $H^a$ stores scores of every actor}}
\For{each $f^a_i$ in $\mathcal{F}^a$}
    \State $\hat{f}^a_i \gets MLP_{\theta_a}(f^a_i)$ 
    \State $h^a_i \gets \mid\mid \hat{f}^a_i \oplus \hat{f}^e \mid\mid_2$   \Comment{\scriptsize{$\oplus$: element-wise addition}}
    \State append $h^a_i$ to $H^a$
\EndFor

\State $H^a \gets softmax(H^a)$ 
\State $\tau \gets \frac{1}{\mid\textbf{h}^a\mid}$ 

\For{each $h^a_i$ in $H^a$}
    \If{$h^a_i > \tau$}
        \State append $f^a_i$ to $\tilde{\mathcal{F}}^a$ 
    \EndIf
\EndFor
\State $f^a \gets self\_attention(\tilde{\mathcal{F}}^a)$ 
\end{algorithmic}
\end{algorithm}

Given $M$ actors (or objects) obtained in the input snippet, only a few of those, i.e., $\hat{M}$ main actors (or relevant objects), actually contribute to an action. Because $\hat{M}$ is unknown and continuously changes throughout the input video, we propose AAM that inherits the merits from adaptive hard attention \cite{adahard_eccv2018} to select an arbitrary number of main actors (or objects) and a soft self-attention mechanism \cite{attention_is_all_you_need} to extract relationships among them. Take actors beholder as an instance, AAM is described by the pseudocode in Algorithm~\ref{algo:aam}.

\subsection{Main Actors - Objects - Environment Interaction}
The third limitation describes situations where the main actors even absent from the scene and only show their hands to perform actions. In these cases, our previous works \cite{KhoaVo_ICASSP, KhoaVo_Access, khoavo_aei_bmvc21} may not work properly due to their reliance on the off-the-shelf actors detector, which can not detect humans for actors modeling. Therefore, we introduce a new entity to comprehensively model the scene in these cases, which is the objects. Capturing objects is very challenging because of two reasons. Firstly, there are various types of objects that can appear in the scenes, and secondly, they frequently appear in very tiny regions, which challenges many existing popular objects detector. To resolve both challenges, we employ the CLIP \cite{radford2021learning}, a powerful pre-trained model that can detect a large amount of objects based on the semantic correlation between their embedding features with the visual features of input image.
Modeling the interactions between three types of entities, i.e., actors, objects, and environment help comprehensively capturing important information for downstream tasks. Our proposed AOE-Net \cite{vo2022aoe} with such modeling method has proved to be very effective in TAPG.

In this section, we would like to detail the last model on AOE as follows:
Given a $N$ frames video $\mathcal{V}=\{v_i\}_{i=1}^{N}$, where $v_i$ is the $i$-th frame, we first follow the standard settings from existing works by segmenting $\mathcal{V}$ into a sequence of $\delta-$frame \textit{snippets} $s_i\mid_{i=1}^T$. Each snippet $s_i$ consists of $\delta$ consecutive frames, therefore, $\mathcal{V}$ has a total of $T=\bigr\lceil \frac{N}{\delta} \bigr\rceil$ snippets. Let $\phi(.)$ be an encoding function to extract the visual feature $f_i$ of a $\delta$-frame snippet $s_i$; the video $\mathcal{V}$ can be represented as $\mathcal{F}$ as follows:
\begin{equation}
        \mathcal{F} =\{f_i\}_{i=1}^{T}, \text{ where } f_i  =\phi(s_i) 
\end{equation}

Different from the existing works \cite{BSN++,dbg, gtan_cvpr2019, xu2020gtad, tsi_accv, bmn, lin2018bsn, xu2020gtad, bai2020boundary, tan2021relaxed}, which simply define $\phi(.)$ as a pre-trained backbone network (e.g., C3D\cite{c3d}, 2Stream \cite{2_stream_1}, SlowFast \cite{SlowFast}), we model $\phi(.)$ by the proposed PMR, which is capable of encoding visual information of multiple entities using both visual and linguistic method.

As stated in Sec.~\ref{sec:intro}, PMR includes four modules, i.e., (i) Environment Beholder, (ii) Actors Beholder, (iii) Objects Beholder, and (iv) Actors-Objects-Environment Beholder. In the sub-sections below, we discuss about each of those modules consecutively, then, we provide details of AAM, which is the main component of Actors Beholder and Objects Beholder to eliminate inessential actors and irrelevant objects, respectively, and extract mutual relationships of main actors and most relevant objects, respectively.

\begin{figure*}[t]
\centering
  \includegraphics[width=\linewidth]{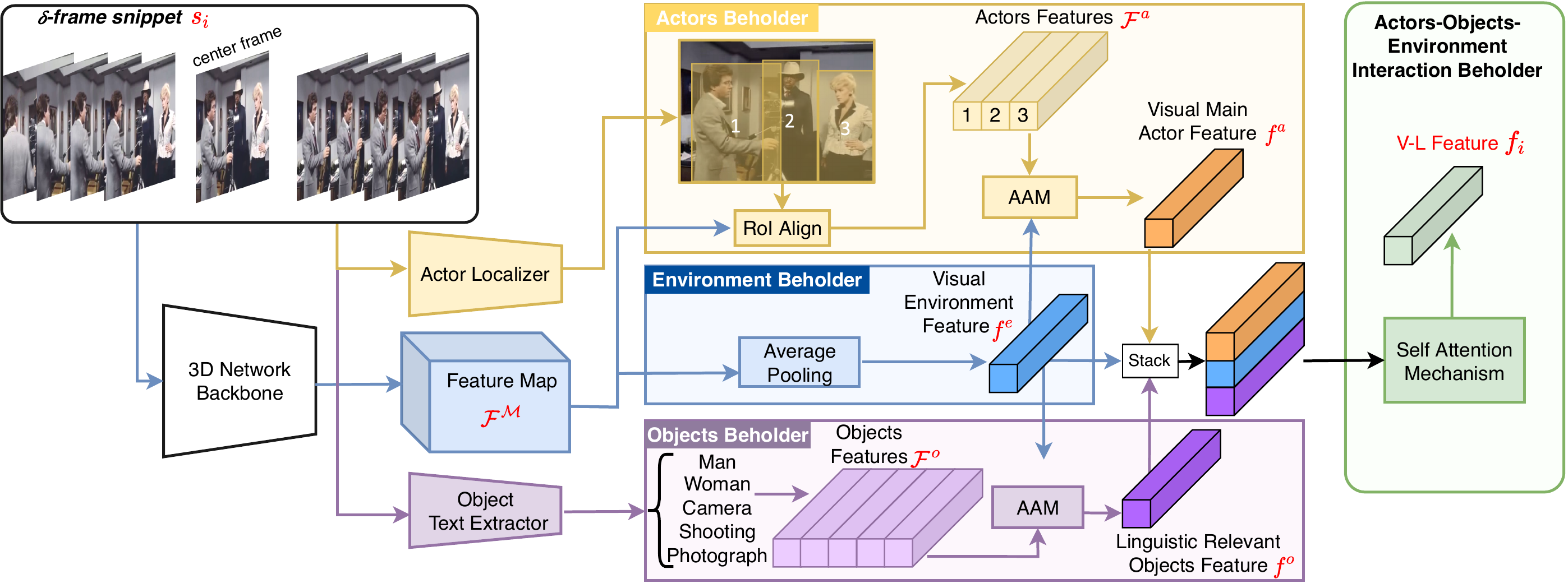}
  \caption{The architecture of PMR. Given a $\delta$-snippet $s_i$, the V-L feature is obtained by four modules: (i) actors beholder to extract local visual action feature $f^a$; (ii) environment beholder to extract global visual environment feature $f^e$; (iii) objects beholder to extract linguistic object feature $f^o$, and (iv) actors-objects-environment interaction beholder to model V-L feature as the interaction between actors, objects and the environment.}
  \label{fig:AOE-PMR}
  \vspace{-5mm}
\end{figure*}

\subsubsection{Environment Beholder}
is responsible for globally capturing visual information of the input $\delta$-frame snippet. To extract both spatial and temporal information of the snippet, we adopt a pre-trained 3D convolutional network as a backbone feature extractor. The snippet is processed through all convolutional blocks of the backbone except the final linear layers to obtain a feature map $\mathcal{F^M}$, then, an average pooling operator is employed to produce an environment feature vector $f^{e}$.

\subsubsection{Actors Beholder}
has a role of semantically extracting visual main actors representation $f^a$. In most cases, an action cannot happen if a human (main actor) is absent notwithstanding environment (Fig.~\ref{fig:motivation}(a)). On the other hand, when an action occurs, it does not necessarily signal that every actor in the scene has committed the action (Fig.~\ref{fig:motivation}(b)). Hence, the Actors Beholder first localizes all existing actors (humans) in a $\delta$-frame snippet by an off-the-shelf object detector onto the middle frame assuming that the actors would not move fast enough to be mis-located with a small $\delta$. We denote $\mathcal{B}=\{b_i\}_{i=1}^{N_B}$ as a set of detected human bounding boxes, where $N_B \geq 0$. Afterwards, each of the detected bounding boxes, $b_i$, is aligned onto feature map $\mathcal{F^M}$ (obtained from Environment Beholder) using RoIAlign \cite{MaskRCNN_ICCV17}. Then, each bounding box feature is average-pooled into a single feature vector $f^a_i$. Finally, we obtain a set of actor features $\mathcal{F}^a=\{f^a_i\}^{N_B}_{i=1}$.

To adaptively select an arbitrary number of main actors and extract their mutual relationships, we apply our proposed AAM, which is explained in Algorithm~\ref{algo:aam}.

\subsubsection{Objects Beholder}
Different from the environment and actors, objects may appear very tiny, in the feature map $\mathcal{F}^\mathcal{M}$. Hence, in this objects beholder, we propose to use linguistic information from relevant objects, which is considerably more informative than visual information. We leverage CLIP~\cite{radford2021learning} as a powerful pre-trained model to extract linguistic information.

As our task just focuses on human activities and their related objects, we utilize the corpus of ActivityNet Captioning annotations \cite{krishna2017dense} to construct the object text vocabulary $\mathcal{T} = \{\mathcal{T}_i\}_{i=1}^{D}$.

ActivityNet Captioning dataset \cite{krishna2017dense} annotates the same set of videos in ActivityNet-1.3 \cite{caba2015activitynet}. Video captions are composed by a vocabulary of up to 10,648 words. In order to create a vocabulary which majorly contains objects and human activities, we eliminate stop words, pronouns, numbers, and infrequent words (which appears 5 times or lower in the whole dataset). Afterwards, we remove words that do not present in the vocabulary of CLIP~\cite{radford2021learning}. To this end, the final vocabulary for our objects beholder consists of $D=3,544$ words.

Each word $\mathcal{T}_i\in\mathcal{T}$ is encoded by a Transformer network \cite{attention_is_all_you_need} into a text feature $\mathcal{T}^f_i$. Let $W_t$ be a text projection matrix pre-trained by CLIP, the embedding text vocabulary is computed as $\mathcal{T}^e = W_t \cdot \mathcal{T}^f$, where $\mathcal{T}^f = \{\mathcal{T}^f_i\}_{i=1}^{D}$. Let $W_i$ be an image projection matrix pre-trained by CLIP, a middle frame $I$ of the $\delta$-frame snippet is first encoded by Vision Transformer \cite{dosovitskiy2020image} to extract visual feature $I^f$, and then embedded by $W_i$, i.e., $I^e = W_i\cdot I^f$. The pairwise cosine similarities between embedded $I^e$ and $\mathcal{T}^e$ is then computed. Top $K$ similarity scores are chosen as output objects text represented by feature $\mathcal{F}^o=\{\mathcal{T}^f_i\}_{i=1}^K$. Similar to the actors beholder, we apply the proposed AAM (described in Algorithm~\ref{algo:aam}) to select relevant objects from $\mathcal{F}^o$, then model the semantic relations among them, and finally obtain linguistic feature $f^o$.

\subsubsection{Actors-Objects-Environment (AOE) Beholder:}
\label{subsubsec:aoe}
AOE Beholder models the relations between global visual environment feature $f^e$, local visual of main actors features $f^a$, and linguistic relevant objects features $f^o$. Firstly, we stack three types of features together as $\mathcal{F}^{aoe}=[f^a,f^o,f^e]$. Then, we employ the self-attention model \cite{attention_is_all_you_need} followed by an average pooling layer to fuse the stack of features $\mathcal{F}^{aoe}$ into $f_i$. $f_i$ is a visual-linguistic feature that represents the input snippet $s_i$ through both visual (environment and actors modalities) and linguistic (objects modality) ways.




\section{Contextual Explanation Representation \\ in TAPG}
\label{subsec:tapg}

To integrate our proposed PMR into TAPG task, we adopt the SOTA method of Boundary Matching Network (BMN)~\cite{bmn} as the action proposals generation module. BMN takes the V-L features sequence $\mathcal{F} = \{ f_i\}_{i=1}^T$ from our PMR as its input. BMN contains three components: semantic modeling, temporal estimation (TE), and proposal estimation (PE). Semantic modeling captures temporal relations between snippets. The TE component evaluates the probabilities of each snippet $s_i\mid_{i=1}^T$ to be an action starting ($P^S_i$) or ending ($P^E_i$) boundaries. Finally, the PE component evaluates every interval $[i,j]$ in the video to estimate its actionness score $P^A_{i,d}$, where $d=j-i$. We refer readers to \cite{bmn, vo2022aoe} for a detailed description on the architecture of BMN.

\subsection{Training Method}
We follow \cite{bmn, lin2018bsn} to generate ground truth labels, including starting labels and ending labels for TE training, and duration labels for PE training.

The starting and ending labels are generated for every snippet of the input video, which are $L^S=\{l^s_n\}_{n=1}^T$ and $L^E=\{l^e_n\}_{n=1}^T$, respectively. A label $l^s_n$ (or $l^e_n$) is set to 1 if its corresponding snippet $s_i$ is the nearest one to any groundtruth starting boundary (or ending boundary).

The duration labels are $L^A \in \{0, 1\}^{D \times T}$ where $D$ is the maximum length of proposals being considered in number of snippets (we set $D=T$ in all of our experiments as suggested in \cite{bmn}). With an element at position $(t_i, t_j)$ stands for a proposal action $a_p=(t_s=\frac{t_j\cdot T}{t_v}, t_e=\frac{(t_j+t_i)\cdot T}{t_v})$, it will be assigned by $1$ if its temporal Interaction-over-Union with any ground truth action in $\mathcal{A}=\{a_i\}_{i=1}^{M}$ reaches a local maximum, or $0$ otherwise.

Three outputs of BMN, i.e., $P^S$, $P^E$, and $P^A$, are trained through three corresponding loss functions of $\mathcal{L}_{s}(P^S, L^S)$, $\mathcal{L}_{e}(P^E, L^E)$, and $\mathcal{L}_{act}(P^A, L^A)$. Where $\mathcal{L}_{s}$ and $\mathcal{L}_{e}$ are defined as weighted binary log-likelihood loss:
\begin{equation*}
\small
    \mathcal{L}_{wb}(P, L) =  \sum^{N}_{i=1}\left[\frac{L_i}{N^+}\log P_i + \frac{(1-L_i)}{N^-}\log (1-P_i)\right]
\label{eq:L_wb}
\end{equation*}
\noindent
where $N^+$ and $N^-$ are the number of positives and negatives in groundtruth labels, respectively. Conversely, $\mathcal{L}_{act}(P,L)$ is defined as follows:
\begin{equation*}
    \mathcal{L}_{act}(P,L)=\mathcal{L}_{wb}(P,L)+\lambda \mathcal{L}_2(P,L)
\end{equation*}
, where $\mathcal{L}_2$ is the mean squared error loss and $\lambda$ is set to $10$.

\section{Contextual Explanation Representation\\ in VPC}
\label{subsec:vpc}

Like TAPG task, in VPC \cite{ dai2019transformer, lei2020mart, yamazaki2022vlcap}, our PMR is employed to extract feature sequences that are served to the Paragraph Generation Module (PGM). PGM operates through each event of the video in the chronological order, then generates a caption describing the event. PGM not only has to maintain the consistency of every word in an event caption, but also need to model the coherency of all captions, to generate a smooth and sound paragraph that describes the input video.

Towards such requirement, we proposed a novel Transformer-in-Transformer (TinT) architecture, which includes (a) an inner Transformer Decoder \cite{attention_is_all_you_need} that generates caption of an event using its corresponding PMR features sequence, and (b) an outer Transformer that maintains the paragraph coherency via self-attention on a set of hidden states, each of which is produced after every event. We refer readers to \cite{kashu_VLTinT} for a detailed description on the process of our TinT method.

\subsection{Training Method}
Given an event $e_k$ and its groundtruth caption $C_k=\{c_i\}_{i=1}^{|C_k|}$, we employ the commonly used Kullback-Leibler (KL) divergence loss as our main training loss $\mathcal{L}_{cap.}$ to train our TinT model so that the predicted caption distribution becomes similar to groundtruth distribution. Besides, following \cite{Song2021} to additionally use a regularization term $\tau(C)$ that penalizes frequently predicted tokens, to reduce redundant phrases in the predicted paragraph. The optimization is illustrated as equations below:
\begin{align*}
\mathcal{L}_{cap.} &= -\frac{1}{N}\sum_{i=1}^{N}(\log{p_\theta(s_i | s_{<i}, \mathcal{V}_e)}) + \lambda\tau(\mathbf{s}) \\
\tau(C) &= -\frac{1}{|C|}\sum_{i=1}^{|C|}\sum_{c\in \{c | C_{<i}\}}\log{(1-p_\theta(c|C_{<i}, \mathcal{E}))}
\end{align*}
where we set $\lambda=0.1$ in all our VPC experiments.

\begin{table}[!tb]
\centering
\begin{tabular}{ll|ccc}
Methods  & Feature  & AR@100 & AUC(val) & AUC(test) \\
\hline\hline
TCN \cite{TCN}            & 2Stream & --      & 59.58    & 61.56     \\
MSRA \cite{MSRA}          & P3D    & --      & 63.12    & 64.18     \\
SSTAD \cite{SSTAD_BMVC17} & C3D     & 73.01  & 64.40    & 64.80     \\
CTAP \cite{CTAP}          & 2Stream & 73.17  & 65.72    & --         \\
BSN   \cite{lin2018bsn}   & 2Stream & 74.16  & 66.17    & 66.26 \\
SRG  \cite{SRG}           & 2Stream & 74.65 & 66.06 & -- \\
MGG \cite{liu2019multi} & I3D & 74.54  & 66.43    & 66.47 \\
BMN   \cite{bmn}      & 2Stream & 75.01  & 67.10    & 67.19 \\
DBG   \cite{dbg}      & 2Stream & 76.65  & 68.23    & 68.57 \\
BSN++   \cite{BSN++}  & 2Stream & 76.52  & 68.26    & -- \\
TSI++ \cite{tsi_accv} & 2Stream & 76.31  & 68.35    & 68.85 \\
MR\cite{MR_eccv2020}  & I3D & 75.27 & 66.51 & -- \\
SSTAP \cite{wang2021self} & I3D & 75.54 & 67.53 & -- \\
TCANet \cite{qing2021temporal} &  2Stream   & 76.08 & 68.08 & -- \\
Zheng, et.al. \cite{zheng2021boundary} & 2Stream & 74.93 & 65.20 & -- \\
\midrule
AEN \cite{KhoaVo_ICASSP}  & C3D & 75.65 & 68.15 & 68.99 \\
ABN \cite{KhoaVo_Access}  &  C3D   & 76.72 & 69.16 & 69.26 \\
AEI \cite{khoavo_aei_bmvc21} & C3D & \underline{\textit{77.24}} & \underline{\textit{69.47}} & \underline{\textit{70.09}} \\
PMR + BMN \cite{vo2022aoe} & C3D & \textbf{77.67} & \textbf{69.71} & \textbf{70.10} \\
\bottomrule
\end{tabular}
\caption{\textbf{TAPG} comparisons on ActivityNet-1.3 \cite{caba2015activitynet} in terms of AR@100 and AUC on validation set and AUC on testing set. Methods in bottom section use the contextual explainable representation as stated in Sec.~\ref{sec:graybox}.}
\label{tab:TAPG-ANet}
 \vspace{-2mm}
\end{table}

\begin{table}[t]
\centering
\resizebox{\linewidth}{!}{
\begin{tabular}{l|l|l|cccc|cc}
Methods & Input & B\@4 $\uparrow$ & M $\uparrow$&  C $\uparrow$ & R $\uparrow$ & Div\@2 $\uparrow$  & R\@4 $\downarrow$\\ \hline \hline
Vanilla Trans. \cite{zhou2018end} & Res200/Flow & 9.31 & 15.54&  21.33& 28.98$^\dag$ &  77.29$^\dag$ & 7.45\\
Trans.-XL \cite{dai2019transformer}  & Res200/Flow & 10.25 & 14.91 & 21.71 & 30.25$^\dag$ & 76.17$^\dag$ &8.79 \\ 
Trans.-XLRG \cite{lei2020mart}  & Res200/Flow & 10.07 & 14.58 & 20.34 & -- & -- &  9.37 \\
MART \cite{lei2020mart} & Res200/Flow &  9.78 & 15.57 & 22.16 & \underline{30.85}$^\dag$ & 75.69$^\dag$ & 5.44 \\
MART$^{\text{COOT}}$ \cite{ging2020coot}  & COOT & 10.85 &  \underline{15.99} & \underline{28.19} & -- & -- &  6.64 \\
Memory Trans. \cite{Song2021}  & I3D &  \underline{11.74} & 15.64 & 26.55 & -- & \textbf{83.95} & \textbf{2.75} \\
\hline
PMR+TinT \cite{kashu_VLTinT} & C3D/Ling & \textbf{14.50} & \textbf{17.97} & \textbf{31.13} & \textbf{36.56} & \underline{77.72} &  \underline{4.75}  \\
\bottomrule
\end{tabular}
}
\caption{Performance comparison of PMR+TinT with other SOTA models on ActivityNet Captions \textit{ae-test}. $\dag$ denotes results obtained by ourselves.} 
\label{tab:anet_test}
\vspace{-5mm}
\end{table}

\section{Experiments}
\subsection{Datasets and Metrics}
For both TAPG and VPC, we evaluate our proposed PMR with BMN \cite{bmn} module on the popular dataset of ActivityNet-1.3, which includes 10,009 training videos, 4917 validation videos, and 5,044 testing videos.

On TAPG, each video is annotated with intervals containing one of 200 activities of interest. We evaluate and compare our method with SOTAs by two common metrics of AR@100 and AUC. AR@100 is the average recall (AR) calculated with an average of 100 proposals per video, while AUC is the area under the AR vs. AN curve score.

On VPC, each video is densely annotated with important events, each event is described by a single sentence. On average, there are 7.7 events per video. Besides, VPC task of ActivityNet-1.3 splits the validation set into ae-val subset with 2460 videos and ae-test subset with 2457 videos. We evaluate and compare our method wtih SOTAs by common metrics in image captioning and video captioning, i.e., BLEU-4 (B@4) \cite{papineni2002bleu}, METEOR (M) \cite{denkowski2014meteor}, and CIDEr (C) \cite{vedantam2015cider}. To evaluate the diversity of generated captions, we use two diversity metrics of 2-gram diversity (Div@2) \cite{div} and 4-gram repetition (R@4) \cite{xiong2018move}.
 
\subsection{Implementation Details}
We employ the C3D \cite{c3d} network pre-trained on Kinetics-400 \cite{Kinetics} as the backbone network in all experiments on both tasks. Features extracted by C3D have 2048 dimensions. 

For Objects Beholder, we adopt the powerful CLIP model \cite{radford2021learning} pre-trained on a large-scale dataset of 400M image-text pairs crawled from the Internet to extract object texts. In the Actors Beholder, to detect humans, we adopt Faster-RCNN model \cite{FasterRCNN} pre-trained on the COCO dataset \cite{cocodataset}. Adam optimizer was used in all experiments, and the initial learning rate is set to 1e-4 for both tasks. 

\begin{figure}[t]
     \centering
     \begin{subfigure}[b]{0.45\textwidth}
         \centering
         \includegraphics[width=\textwidth]{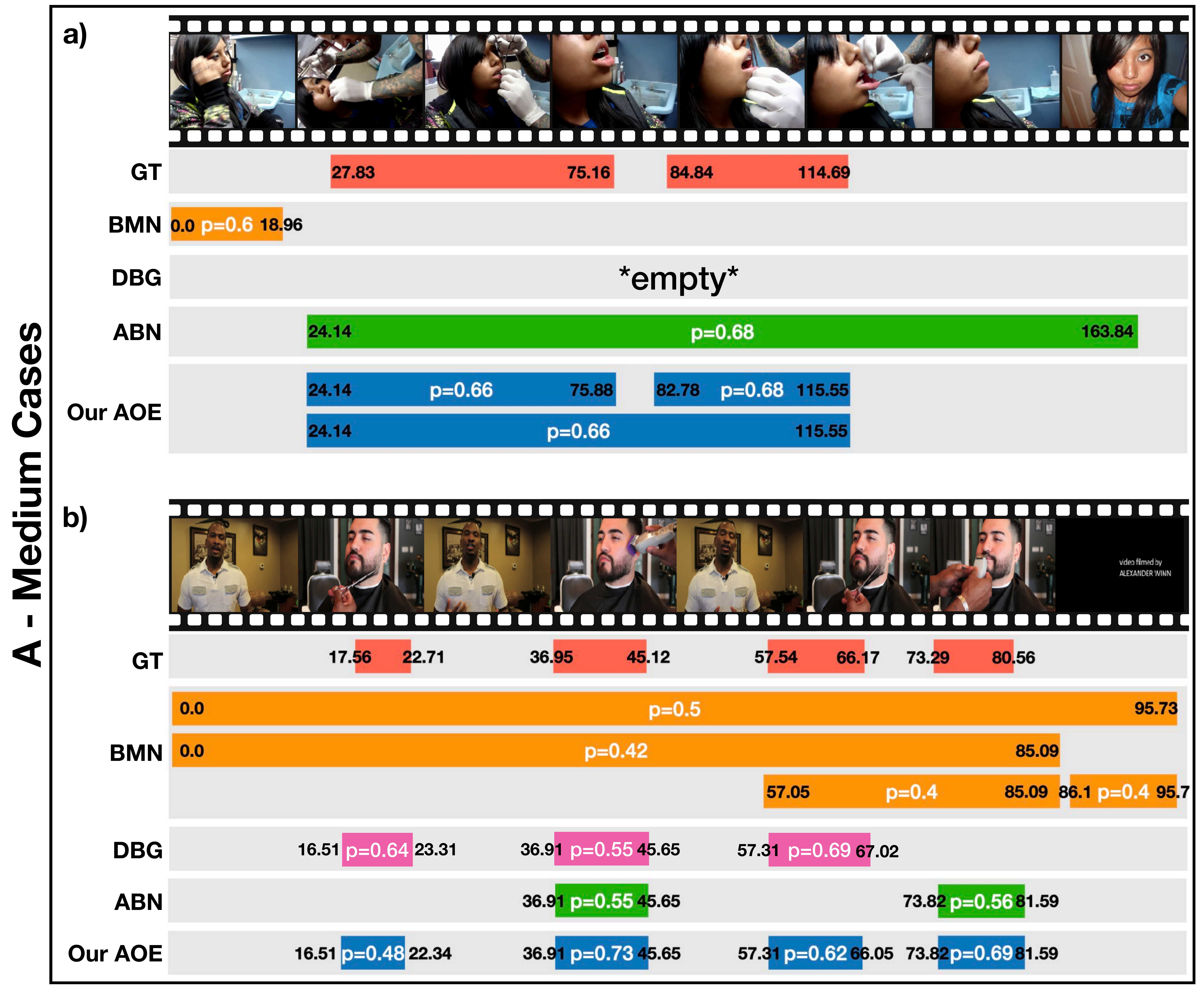}
         \label{fig:anet2}
     \vspace{-3mm}
     \end{subfigure}
     \begin{subfigure}[b]{0.45\textwidth}
         \centering
         \includegraphics[width=\textwidth]{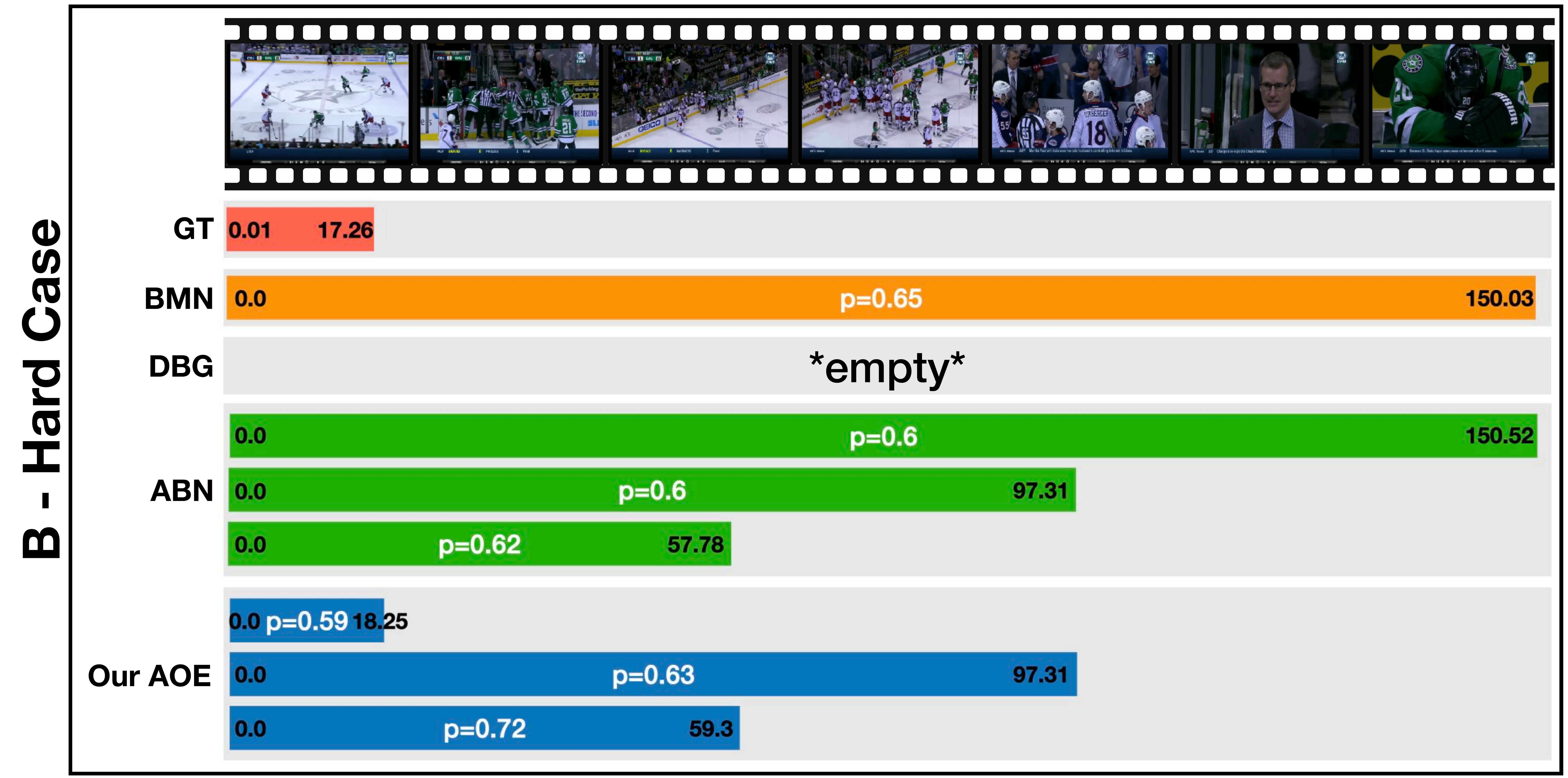}
         \label{fig:anet3}
     \vspace{-3mm}
     \end{subfigure}
     \begin{subfigure}[b]{0.45\textwidth}
         \centering
         \includegraphics[width=\textwidth]{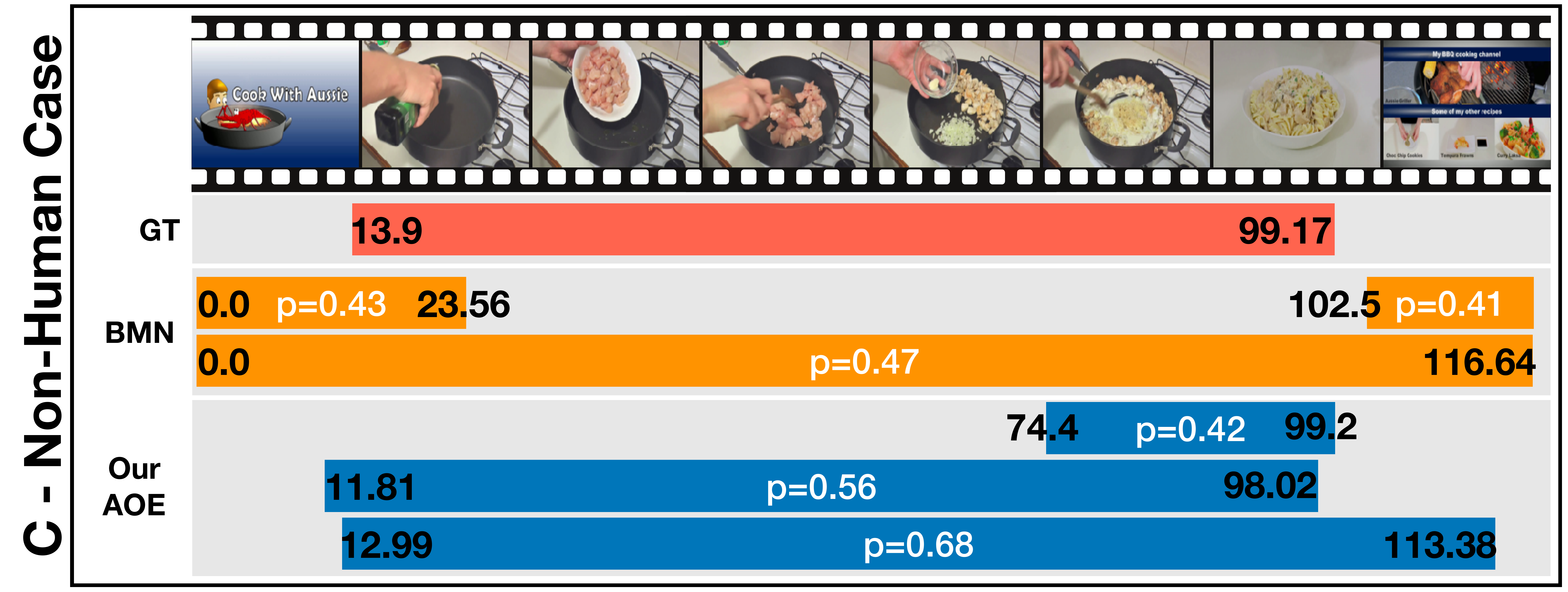}
         \label{fig:anet4}
     \end{subfigure}
\vspace{-5mm}   
\caption{Qualitative results in TAPG on ActivityNet-1.3 \cite{caba2015activitynet} dataset.}
\label{fig:qualitative_anet}
\vspace{-5mm}   
\end{figure}

\begin{table}[!t]
\centering
\resizebox{\linewidth}{!}{
\begin{tabular}{l|ccccc|c|c|c|c|c}
\multirow{2}{*}{\textbf{Exp}} & \multicolumn{5}{c|}{\textbf{Setting}} & \multicolumn{5}{c}{\textbf{TAPG Performance}} \\
& Act. & Env. & Obj. & AAM & Soft-Att & @50 & @100 & @200 & @500 & @1000 \\
\hline\hline
\#1 & \colorbox{cyan}{\cmark} & \xmark & \xmark & \xmark &\colorbox{cyan}{\cmark} & 25.96 & 35.14 & 43.48 & 52.37 & 58.47 \\
\#2& \xmark & \colorbox{cyan}{\cmark} & \xmark & \xmark & \xmark & 38.94 & 47.80 & 54.93 & 61.92 & 65.96 \\
\#3& \xmark & \xmark & \colorbox{cyan}{\cmark} & \xmark &\colorbox{cyan}{\cmark} & 18.06 & 26.68 & 37.14 & 49.28 & 56.99 \\
\hline
\#4& \colorbox{cyan}{\cmark} & \colorbox{cyan}{\cmark} & \xmark & \xmark &\colorbox{cyan}{\cmark} & 40.87 & 49.09 & 56.24 & 63.53 & 67.29 \\
\#5& \colorbox{cyan}{\cmark} & \colorbox{cyan}{\cmark} & \colorbox{cyan}{\cmark} & \xmark &\colorbox{cyan}{\cmark} & 42.60 & 49.86 & 56.87 & 63.76 & 67.60 \\
\#6& \colorbox{cyan}{\cmark} & \colorbox{cyan}{\cmark} & \xmark & \colorbox{cyan}{\cmark} & \xmark & 43.79 & 49.67 & 56.73 & 63.49 & 67.36 \\ \hline
\#7& \colorbox{cyan}{\cmark} & \colorbox{cyan}{\cmark} & \colorbox{cyan}{\cmark} & \colorbox{cyan}{\cmark} & \xmark & 44.56 & 50.26 & 57.30 & 64.32 & 68.19 \\ \bottomrule
\end{tabular}
}
\caption{TAPG comparisons on different network settings. Act., Env., Obj. denote actors, environment, objects beholders.} 
\label{tb:abl_1}
\vspace{-5mm}   
\end{table}

\begin{table*}[!hbt]
\centering
\resizebox{0.95\linewidth}{!}{
\begin{tabular}{ccc|cccc|cc||cccc|cc}
\toprule
& & & \multicolumn{6}{c||}{\textit{at-test} split} & \multicolumn{6}{c}{\textit{ae-val} split}\\ \hline
Env. & Act. & Obj. & B@4 $\uparrow$ & M $\uparrow$&  C $\uparrow$ & R $\uparrow$ &  Div@2$\uparrow$ & R@4 $\downarrow$ & B@4 $\uparrow$ & M $\uparrow$&  C $\uparrow$ & R $\uparrow$ & Div@2 $\uparrow$ & R@4 $\downarrow$ \\ \hline
\colorbox{cyan}{\cmark} & \xmark & \xmark & 13.62 & 17.41 & 29.09 & 35.96 & 76.14 & 5.97 & 14.02 & 17.58 & 30.31 & 36.20 & 76.11 &6.08  \\
\xmark & \colorbox{cyan}{\cmark} & \xmark & 11.83 & 16.22 & 21.39 & 33.97 & \underline{79.20} & \underline{4.16} & 12.13 & 16.57  & 24.98  & 34.36 & \underline{79.18}  & \underline{4.24}\\
\xmark & \xmark & \colorbox{cyan}{\cmark} & 13.38 & 17.69 & 30.30 & 35.63 & \textbf{80.50} & \textbf{3.32}  & 14.00 & 17.88  & 31.64  & 35.95 & \textbf{80.44}  & \textbf{3.22}\\ 
\colorbox{cyan}{\cmark} & \colorbox{cyan}{\cmark} & \xmark &  13.77 & 17.52 & 30.05 & 35.93 & 77.78 & 4.69 & 14.12 & 17.78  & 31.15  & 36.12 & 78.02  & 4.56\\ 
\colorbox{cyan}{\cmark} & \xmark & \colorbox{cyan}{\cmark} & \textbf{14.53} & \underline{17.79} & \underline{30.83} & \textbf{36.67} & 76.47 & 5.60 &
\underline{14.84} & \underline{17.97} & \underline{31.86} & \underline{36.80} & 76.41 & 5.67 \\ 
\colorbox{cyan}{\cmark} & \colorbox{cyan}{\cmark} & \colorbox{cyan}{\cmark} & \underline{14.50} & \textbf{17.97} & \textbf{31.13} & \underline{36.56} & 77.72 & 4.75 & \textbf{14.93} & \textbf{18.16}  & \textbf{33.07}  & \textbf{36.86} & 77.72  & 4.87\\ \bottomrule
\end{tabular}
}
\caption{VPC comparisons on different network settings. Env., Act., and Obj. denote the global visual environment, local visual main agents, and linguistic relevant objects, respectively.}
\label{tab:ablation_feat}
\vspace{-5mm}
\end{table*}

\begin{figure*}[!hb]
\vspace{-5mm}
    \centering
    \includegraphics[width=0.9\linewidth]{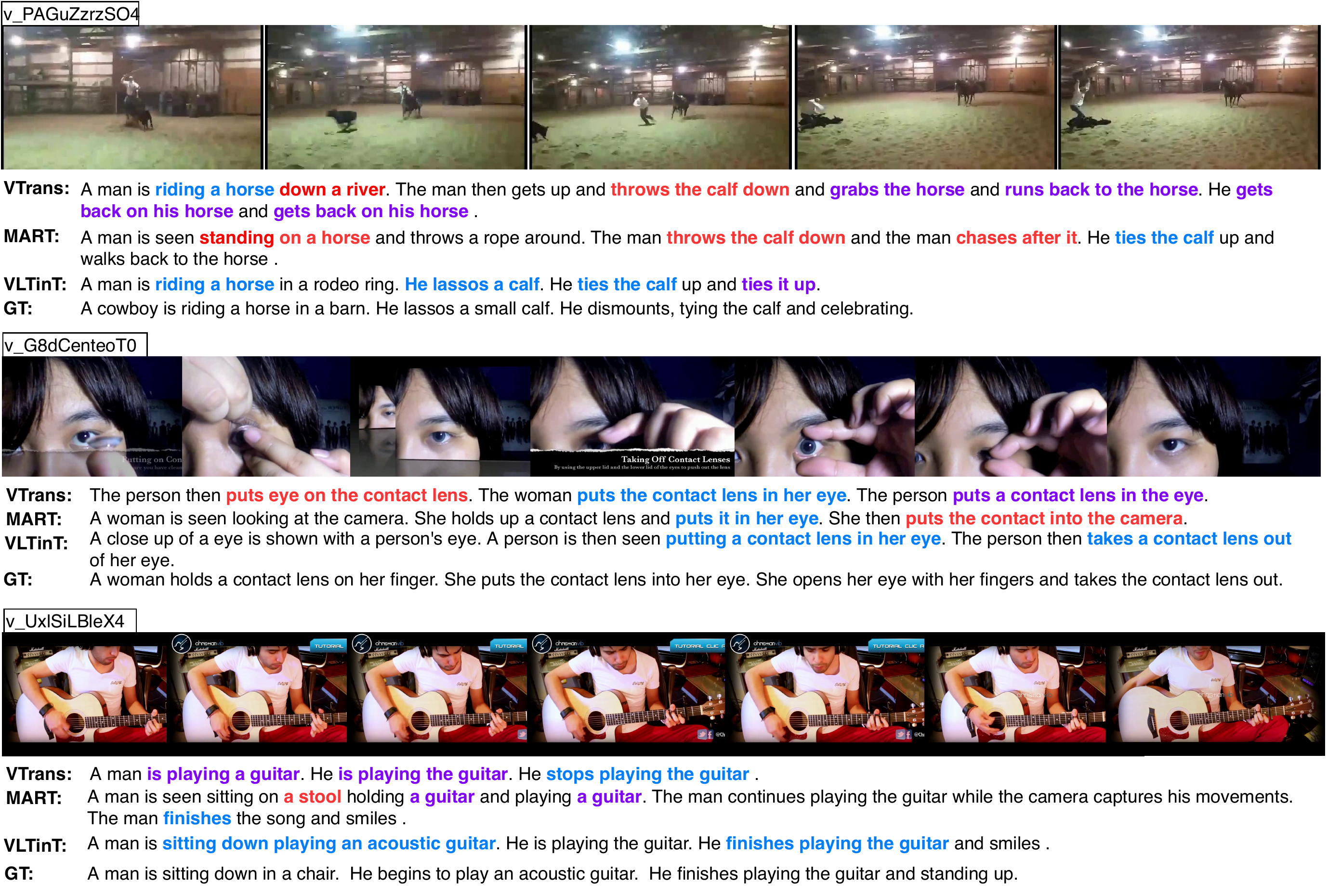}
    \caption{Qualitative comparison on ActivityNet Captions \textit{ae-test} split between our VLTinT and VTrans\cite{zhou2018end}, MART \cite{lei2020mart}. At each video, captioning from VTrans is in the $1^{st}$ row, MART is in the $2^{nd}$ row, our VLTinT is in the $3^{rd}$ row, and groundtruth (GT) is in the $4^{th}$ row.
    \textcolor{red}{Red text} indicates the captioning mistakes, \textcolor{violet}{purple text} indicates repetitive patterns, and \textcolor{blue}{blue text} indicates some distinct expressions.
    We compared our model with Vanilla Transformer (VTrans) and MART as baselines. GT indicates the groundtruth captioning.}
    \label{fig:qualitative}
\end{figure*}

\subsection{Performance and comparison on TAPG}
Table \ref{tab:TAPG-ANet} presents the evaluation of our PMR on TAPG and comparisons with previous SOTAs on ActivityNet-1.3 \cite{caba2015activitynet}. The experimental results demonstrate that our proposed representation with BMN outperforms the existing methods in terms of AR@100 and AUC by an adequate margin. Notably, the performance on TAPG of our AOE-Net is competitive with AEI-B \cite{khoavo_aei_bmvc21}, which is followed closely by ABN \cite{KhoaVo_Access}, both of which also incorporate local actors and global environment. This experiment strongly supports our observation and motivation on using the human perception principle to analyze human actions in untrimmed videos.

\subsection{Performance and comparison on VPC}
We benchmark and compare our PMR and TinT modules on VPC task with the prior SOTAs on both ActivityNet Captions \textit{ae-test} in Table \ref{tab:anet_test}. Compared to SOTA approaches, i.e., MART \cite{lei2020mart}, MART w/COOT \cite{ging2020coot}, and PDVC \cite{wang2021end}, our approach outperforms with large margins on both accuracy and diversity metrics on ActivityNet Captions. For example, the accuracy gains 3.65\%/1.98\%/2.94\%5.71\% on B@4/M/C/R metrics whereas diversity increases 0.43\% on Div@2 and reduces 0.67\% on R@4 compared to the second-best performance. Qualitative comparisons on VPC are illustrated in Fig.~\ref{fig:qualitative}.


\section{Conclusion}
In this paper, we present a novel video representation method, namely Perception-based Multi-modal Representation (PMR), which simulates the human perception process. Our PMR extracts the visual-linguistic representation of each snippet with four modules. Environment beholder and actors beholder capture global and local visual features of environment and main actors, respectively. Objects beholder extracts linguistic feature from relevant objects. The last beholder aims to model the relations between main actors, relevant objects and environment. To focus on an arbitrary number of main actor(s) or relevant objects, we introduced AAM. 

We evaluate PMR on two untrimmed videos understanding tasks, i.e., temporal action proposals generation (TAPG) and video paragraph captioning (VPC). On TAPG, we employ the SOTA method of BMN~\cite{bmn} as the proposals generation module, while on VPC, we propose a novel Transformer-in-Transformer architecture~\cite{kashu_VLTinT} as the paragraph generator. On both tasks, we reported the quantitative and qualitative results, which suggest that our proposed PMR makes an adequate improvement to the selected SOTA modules.

However, we also observe several limitations, which shows some room for further research to improve our PMR. First, the Objects Beholder only represents objects as text features, however, the visual appearance and motions introduced to those objects may be good information to the representation. Second, the Actors Beholder assume humans as actors, but in a general scenario, actors can also be animals, therefore, it is more beneficial if Actors Beholder can learn to localize actors instead of relying on an off-the-shelf objects detector.

\vspace{4mm}
\textbf{Acknowledgments:}
This material is based upon work supported by the National Science Foundation (NSF) under Award No OIA-1946391, NSF 1920920, NSF FAIN-2223793 and NIH 1R01CA277739.

\newpage
\bibliography{main}
\bibliographystyle{IEEEtran}

\end{document}